# Physical Exercise Recommendation and Success Prediction Using Interconnected Recurrent Neural Networks

Arash Golibagh Mahyari

Florida Institute for Human and Machine Cognition (IHMC), 40 S Alcaniz Ave, Pensacola, FL 32502, amahyari@ihmc.us

Peter Pirolli

Florida Institute for Human and Machine Cognition (IHMC), 40 S Alcaniz Ave, Pensacola, FL 32502, ppirolli@ihmc.us

**ABSTRACT**

Unhealthy behaviors, e.g., physical inactivity and unhealthful food choice, are the primary healthcare cost drivers in developed countries. Pervasive computational, sensing, and communication technology provided by smartphones and smartwatches have made it possible to support individuals in their everyday lives to develop healthier lifestyles. In this paper, we propose an exercise recommendation system that also predicts individual success rates. The system, consisting of two inter-connected recurrent neural networks (RNNs), uses the history of workouts to recommend the next workout activity for each individual. The system then predicts the probability of successful completion of the predicted activity by the individual. The prediction accuracy of this interconnected-RNN model is assessed on previously published data from a four-week mobile health experiment and is shown to improve upon previous predictions from a computational cognitive model.

## 1. Introduction

A major driver of healthcare costs in the U.S. and elsewhere are unhealthy behaviors such as physical inactivity, increased food intake, and unhealthful food choice [1, 2]. Behavioral and environmental health factors account for more deaths than genetics [1]. With the advent of pervasive computational, sensing, and communication technology, there is an opportunity to support individuals in their everyday lives to develop healthier lifestyles. For instance, the pervasive use of smartphones is a potential platform for the delivery of behavior-change methods at great economies of scale. Commercial systems such as noom [3] aim to provide psychological support via mobile health (mHealth) systems. Research platforms, such as the Fittle+ system [4], have demonstrated the efficacy of translating known behavior-change techniques [5] into personal mHealth applications. Computational cognitive models have successfully simulated the day-by-day individual-level effects of such interventions in Fittle+ experiments [6, 7]. However, they fall short in recommending personalized health goals to individuals. In this paper, we develop a recurrent deep neural network architecture to recommend specific health goals to specific individuals such that there is a high probability of achieving those goals. The goal of this novel recommender is not just to recommend exercises, but exercises that individuals have a high likelihood of performing. An exercise recommender is essentially learning to predict the choices (to perform daily exercises or not) that depend largely on individual psychology.

Several psychological theories suggest that an ideal automated mHealth goal recommendation algorithm should modulate the perceived difficulty of goals for each individual [8, 9, 10, 11]. Behavior-change goals are typically defined to be specific and measurable. Specific means unambiguous descriptions of behavior ("drink five glasses of water"), and measurable means reasonably precise evaluation of success or failure. The perceived difficulty of a health goal depends on each individual's history. Goal Setting Theory [8] predicts that goals need to be challenging enough to be motivating. The Social Cognition Theory of self-efficacy [9, 10] predicts that goals that are perceived as too difficult are unlikely to be attempted. Social Cognitive Theory also predicts that self-efficacy can be increased through a process whereby users succeed at progressively more difficult or complex goals over time (guided mastery). Greater levels of self-efficacy lead to greater likelihoods of achieving a goal. The Attributional Theory of Performance [11] proposes that the level of intended effort motivating a performance will increase with the difference between self-efficacy and the perceived difficulty of achieving a goal. Even though there have been several physical activity recommendation systems, they have failed to consider the probability of success for recommended exercises. Recommended behavior goals for individual users should be difficult enough to be motivating, but easy enough to be successfully achieved.

In this paper, we develop a novel recommendation system that recommends exercise goals with high probability of their successful completion by users. To do this, we build upon (1) a 28-day experimental study of a mobile health system called DStress [12] that evaluated a hand-engineered adaptive exercise recommender, and (2) an explanatory computational cognitive model, based on theories described above, fit post-hoc to the DStress data to predict every individual's observed success likelihood for every recommended exercise every day over 28 days [4]. The novel recommender system we described here, trained with user data: (1) recommends exercises in a way that adapts to individual users without prior hand-engineering and (2) more accurately predicts exercise success rates than the previous computational cognitive model.

We use these DStress data to train a recommendation system that uses a combination of association rules and recurrent neural networks. The aim of this technique is to accurately recommend exercises that have a high probability of being pursued and achieved, for each individual on each day. The remaining of this paper is organized as follows: Section 2 is devoted to the related work. The proposed recommendation system is described in Section 3. Section 4 is devoted to the experimental results of the recommendation system and its comparison with the predictions to the computational cognitive model previously developed for the DStress data.

## 2. Related Work

Recommendation systems have been used in online shopping for several years. The goal of recommendation systems is to recommend products that suit the consumers' tastes. Traditional recommendation systems have used collaborative filters to suggest products similar to those the consumers have purchased. With the advancements in deep learning algorithms, several studies have proposed deep learning-based recommendation systems [13]. In [13], a multi-stack recurrent neural network (RNN) architecture is used to develop a recommendation system to suggest businesses in Yelp based on their reviews. Wu et al. [14] used an RNN with long-short term memories (LSTM) to predict future behavioral trajectories. A few studies have proposed exercise recommendation systems [15, 16]. Sami et al. [15] used several independent variables recommends various sports such as swimming using collaborative approaches. Ni et al.[17] developed an LSTM-based model called FitRec for estimating a user's heart rate profile over candidate activities and then predicting activities on that basis. The model was tested against 250 thousand workout records with associated sensor measurements including heart rate.

### DStress Finite State Machine Recommender Study

The data we use comes from the Konrad et al. [12] mHealth experiment with DStress. It was developed to provide coaching on exercise and meditation goals for adults seeking to reduce stress. The purpose of the experiment was to test the efficacy of an adaptive daily exercise recommender (DStress-adaptive) against two alternative exercises programs in which the daily exercises changed according to fixed schedules (Easy-fixed and Difficult-fixed). The DStress-adpative recommender was a hand engineered finite state machine. The transition rules are described in more detail in Konrad et al. [12], but they implement a policy whereby, if a person successfully completes all three exercises assigned for a day, they advance to the next higher level of exercise difficulty. If they do not succeed at exercises or meditation activities, then they are regressed to exercises or meditation activities at an easier level of difficulty. The 46 exercises used in DStree and their difficulty ratings were obtained from three certified personal trainers (e.g., Wall Pushups, Standing Knee Lifts, Squats, and Burpees, etc.). The experiment took place over a 28-day period. In a given week, users encountered three kinds of days: Exercise Days (occurring on Mondays, Wednesdays, Fridays), Meditation Days (Tuesdays, Thursdays, Saturdays), and Rest Days (Sundays).

Adult participants (19-59 yr) were randomly assigned to three conditions with different 28-day goal progressions: (1) a DStress-adaptive condition using the adaptive coaching system in which goal difficulties adjusted to the user based on past performance, (2) an Easy-fixed condition in which the difficulty of daily goals increased at the same slow rate for all participants assigned to that condition, and (3) a Difficult-fixed condition in which the goal difficulties increased at a greater rate. Konrad et al. [12] found that the adaptive DStress-adaptive condition produced significant reductions in self-reported stress levels compared to the Easy-fixed and Difficult-fixed goal schedules. The DStress-adaptive condition also produced superior rates of performing assigned daily exercise goals.

### Explanatory Cognitive Model Prediction of DStress Exercise Success Data

The individual-level day-by-day success rate data were modelled [5] using the ACT-R neurocognitive architecture [3, 6]. ACT-R is a computational architecture for simulating and understanding human learning and cognition. The ACT-R model implemented the psychological theories discussed above and the simulations were used to predict each individual's success in performing each assigned exercise in the 28-day DStress dataset. Specifically, the ACT-R computational model implemented self-efficacy processes that predict that exercise goals that are perceived as too difficult in comparison to perceptions of one's own abilities are unlikely to be attempted, and the Attributional Theory of Performance that proposes that the level of intended effort motivating a performance will increase with the difference between self-efficacy and the perceived difficulty of achieving a goal. The ACT-R model assumes that self-efficacy and intended effort are fundamentally the result of memory processes. Past experiences of efficacy at behaviors similar to a target goal are retrieved from memory and blended together to produce assessments of self-efficacy and intended effort for the new goal. Consequently, the dynamics of self-efficacy and performance exhibit the dynamics of the underlying memory mechanisms, and exhibit well-known memory phenomena. The ACT-R mechanisms and implementation of these psychological processes accurately predict the observed DStress day-by-day individual exercise success, and provide a benchmark for our new exercise recommender systems

## 3. Proposed Algorithm

Our proposed method is an innovative architecture of two inter-connected RNNs that takes exercise embedding and exogenous data as the input and recommends exercise activity and its probability of successful completion in the output.

### 3.1. Exercise Embedding

In the proposed work, the exercise names are converted to vector representations by embedding the names onto a vector space. Word2vec is the most well-known word embedding set of algorithms. Word2vec methods state that words that appear in the same contexts (a window) tend to purport similar meanings [18], and finds low-dimensional distributional word representations. Word2vec methods have been applied to recommendation systems previously [18]. In this paper, we use SkipGram [19] to find the vector representation of exercise names. SkipGram trains a neural network to find the vector representation based on the context (words within a window around the current word). The sequence of exercises performed by one individual in the training dataset is considered as one sentence for the word2vec algorithm. However, sequences of exercises have a special structure induced by the user's behaviour and the exercise' characteristics [18]. Additionally, it is more difficult and challenging to develop accurate embedding algorithms for recommendation systems because the performance of Word2vec algorithms, e.g., SkipGram, is affected by top exercises often representing most of the content users select [18]. The second challenge with sequences of exercises for different individuals is that people with the same exercises on the first day may choose different exercises on the second day. For example, in Figure 1, individuals who have received the recommended exercises "Two Leg Hip Bridge on Floor" And "Pushups off Wall" selected different exercises such as Crunches and Deadbug afterward. This causes an inaccurate exercise prediction because three words "*Two Leg Hip Bridge on Floor*", "*Pushups off Wall*", and Crunches do not appear frequently together instead other variants such as "*Two Leg Hip Bridge on Floor*", "*Pushups off Wall*", and "*Deadbug*" appears in other sequences. To mitigate this problem, we propose to embed the set of exercises that come as descendants close to each other in the vector representation space. The idea is not to replace all exercise names in this set with one exercise, but to improve the accuracy of the predictive model, especially when the training data is small. The preprocessing step is not to eliminate variability in the data. To find the set of exercises that appear together—e.g., {*Crunches, Deadbug, Dumbbell Kneeling Row, Kneeling Plank, Marching in Place, Modified Burpee No Jump, Pushups on Bar, Standing Knee Lifts, Static Lunge With Wall*} in Figure 1—an association analysis algorithm is applied to the dataset [20]. Association analysis discovers interesting relationships in large data, and represents them as association rules or set of frequent items. Figure 1 shows an example of association rules extracted from an exercise dataset. We explain the justification of using SkipGram within the context of Fig1. After embedding the exercise names using SkipGram, the SkipGram model assigns vector representations close to each other for {*Crunches, Deadbug, Dumbbell Kneeling Row, Kneeling Plank, Marching in Place, Modified*

*Burpee No Jump, Pushups on Bar, Standing Knee Lifts, Static Lunge With Wall*}. When the predictive model, e.g., RNN, predicts a vector representation as the next exercise for the sequence of {"*Two Leg Hip Bridge on Floor*", "*Pushups off Wall*"}. The SkipGram, then, converts the vector representation back to the exercise names. Because the set of common exercises coming after {"*Two Leg Hip Bridge on Floor*", "*Pushups off Wall*"} are close to each other, SkipGram will select an exercise name among the set {*Crunches, Deadbug, Dumbbell Kneeling Row, Kneeling Plank, Marching in Place, Modified Burpee No Jump, Pushups on Bar, Standing Knee Lifts, Static Lunge With Wall*} that is the closest to the predicted vector.

After extracting association rules, the set of all exercises appear as consequences of a sequence of exercises that are considered as one sentence for the word2vec algorithm. The reason for choosing the set of consequences as a sentence is that these exercises appear interchangeably so they should be embedded close to each other in the vector representation space.

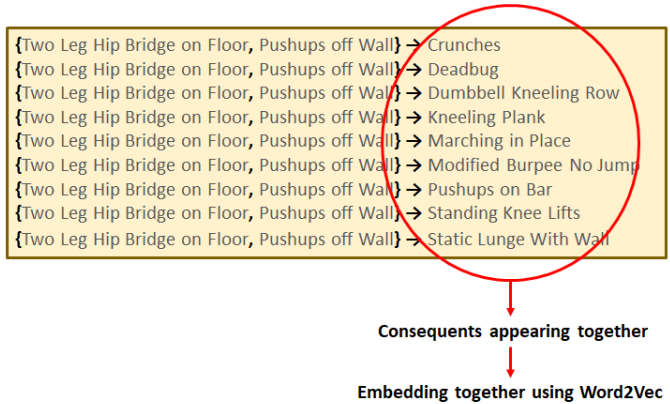

**Figure 1. Set of exercises with the same antecedent exercises within a window.**

## 3.2. Exercise and Its Success Probability Prediction

Our goal is to recommend an exercise to an individual based on their previous exercises and their success rate. This problem is formulated as a time series and success rate prediction. Let's assume that we represent the vector representation of the $i$th exercise with $v_i \in R^N$, where $N$ is the dimension of the vector representation. For each individual, we have a sequence of training dataset $\{(v_1, x_1, p_1), (v_2, x_2, p_2), …, (v_K, x_K, p_K)\}$, where $K$ is the total length of the sequence, pi is the indication of successful completion of the exercise with values $\{0, 1\}$, and $x_i \in R^J$ is the $J$-dimensional exogenous data of the $i$th exercise. The exogenous data are the additional information that we have about each exercise, such as its difficulty level and the body parts involved in the exercise. The successful completion of an exercise depends on various factors, including the previous exercises an individual has received, as well as information about the exercise. For example, the probability of successful completion of the exercise depends on its difficulty, the body part it targets, and the similar information of previous exercises.

This paper used two interconnected RNNs for exercise recommendation and success rate prediction. Figure 2 shows the architecture of the proposed inter-connected RNNs. The first RNN is trained to predict the next exercise according to the history of exercises. The RNN network will be trained on the sequence $\{v_1, v_2, …, v_K\}$. The network will predict $v_i$ based on $\{v_{i-w}, v_{i-w}, …, v_{i-1}\}$, $w$ is the window length. We examined the sequence dependency of the training data using statistical time series methods, e.g., autocorrelation, to select the appropriate window length. We propose to use RNNs if the temporal dependency is small—such as our dataset and to use RNNs with LSTM units when the temporal dependencies are long. Since there is no long-term dependency among the sequence of our exercise data, we have selected a two-layer RNN with $N$ nodes for this purpose. The cost function selected to predict the next exercise is mean square error (MSE).

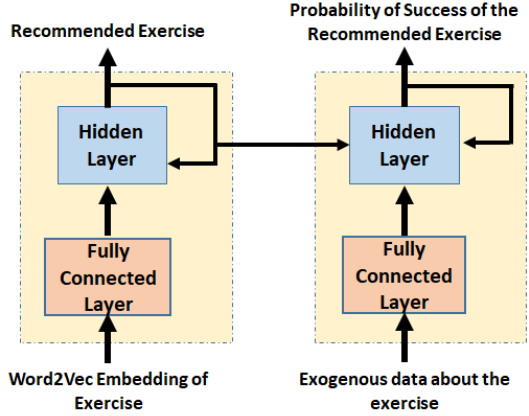

**Figure 2: The architecture of RNN for predicting exercises and their probability of success.**

The second RNN architecture is designed to predict the probability of the recommended exercise. The architecture is inspired by ARMAX model. The probability of success depends on the predicted exercise as well as previous exercises. The probability of success is also affected by exogenous data. We use a two-layer RNN architecture with $N + J$ nodes. The network gets the exogenous data of each predicted exercise as well as the latent representation of exercises from the hidden layer of the first RNN. The second RNN learns to predict $P(p_i | x_i, \hat{v}_i, x_{i-1}, v_{i-1}, \ldots, x_{i-w}, v_{i-w})$. The below equation represents the proposed model in the form of time series analysis.

$$p_i = f_0(\rho \hat{\tau}_i) + f_1\left(\sum_{j=i-w}^{i-1} \alpha_{j-i+w} p_j\right) + f_2\left(\sum_{j=i-w}^{i-1} \beta_{j-i+w} \tau_j\right) + f_3\left(\sum_{j=i-w}^{i} \gamma_{j-i+w} x_j\right)$$

where $f_l$s are nonlinear functions—ReLu and Sigmoid, $\alpha_j, \beta_j, \gamma_j$ are weights that will be learned from the training data, $\tau_j$ is the latent representations of $v_j$ obtained from the hidden layer of the first RNN, $\hat{\tau}_i$ is the latent presentation of the predicted exercise, and $\rho$ is its coefficients to be learned from the training data. The cost function used for training the second RNN is the binary cross-entropy.

## 4. Experimental Results
### 4.1. Data

This paper uses DSTRESS dataset, collected through a mobile application. DStress [11] is a web- and mobile-based system that provides automated coaching on exercise and meditation goals aimed at reducing perceived stress. As discussed above, An experimental evaluation of this app [11] took place over 28 days with interleaved Exercise Days with Meditation Days and one Rest Day per week. The experiment compared three groups of adult participants: (1) a DStress-adaptive group who used the adaptive coaching system in which goal difficulties adjusted to the user based on past performance, (2) an Easy-fixed group in which the difficulty of daily goals increased at the same slow rate for all and (3) a Difficult-fixed group in which the goal difficulties increased at a greater rate. The participants in the DStress-adaptive group self-reported lower stress levels compared to the Easy-fixed and Difficult-fixed groups. The DStress-adaptive group reported a higher level of activity completions than the other two groups (Figure 3) even though they were performing more difficult exercises than the Easy-fixed group. This increased ability to tackle more difficult goals is consistent with a build-up in self-efficacy. We compare our RNN-based approach to the ACT-R model-fits below.

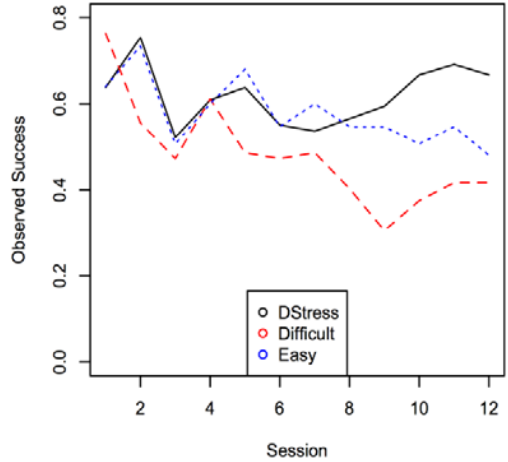

**Figure 3- Summary data from the DStress experiment [11]: the mean rate of participants successfully completing assigned exercises per session over 28 days (3 times per week).**

There were 72 participants in this study. Participants were assigned different exercises of differing difficulty levels depending on experimental condition and personal history. After each exercise, participants indicated whether they have finished the exercise successfully or not. The exercise difficulty levels were rated by three experts. Moreover, the dataset includes lag (days since last successful performance of the exercise) and frequency (cumulative successful performances of the exercise) information for each exercise. Each participant performed 36 exercises.

## 4.2. Experiment Setup

We have used the k-folding approach for training and testing our algorithm, where $k$ is 72. At each time, we leave out one participant from the training and train the architecture on the remaining 71 participants' data. Then, we evaluate our algorithm on one test participant. The statistical analysis of the exercise time series has determined the length of the window equal to $w = 3$. This means that the temporal dependencies among the exercises is at most three-time steps. This has created 2650 sequences of length 3. The exercises are embedded onto a space of dimension $N = 20$ using word2vec algorithm. The exogenous data are *Lag*, *Frequency*, and *Difficulty Level* ($K = 3$). The adaptive learning optimization algorithm "Adam" is used to train the RNN modules [21]. The learning rate is $lr = 0.01$.

Two RNNs are interconnected, so the inference is performed at the same time. However, we will describe their evaluations separately. Two approaches are selected for evaluating the first RNN—exercise prediction—because the first RNN gets its input from and gives its output to Word2Vec. Thus, we evaluate the RNN accuracy on the training and test data using MSE (mean square error). We also compare the *top5* predicted exercises with the list of co-embedded exercises. Because several exercises appear together as consequences of a given sequence, if our proposed algorithm predicts any of the consequences, we call it a success. We explained how we co-embed these consequences together using Word2Vec approach in §3.1. We use the same approach to find the co-embedded list of exercises. If the co-embedded list and the top5 predicted list have common names, we consider it as a true positive (TP). If there is no overlap, it is a false negative (FN). We use this approach to calculate the precision and recall for the first RNN. Figure 4 shows the test accuracy of the first RNN for predicting the next exercise. Figure 4a shows the histogram of the test accuracy of the exercise name prediction for subject $i$ when subject $i$ is left out during the training. To calculate the accuracy, we have left out subject $i$ and trained the model on the remaining subjects. Then, the test accuracy of exericse prediction is calculated for subject $i$ as described above with top5 prediction. In Figure 4a, the accuracy of exercise prediction is on average %80. The accuracy includes the prediction of exercise vector representations and the exercise prediction names from the predicted vector representations. Figure 4b shows the histogram of the mean square error (MSE) for the same experiment. MSE is computed between the activity embedding $v_i$ and the predicted vector representation $\hat{v}_i$. In Figure 4b, the MSE values are in the order of $1e^{-05}$ that shows the great performance of the trained RNN for predicting the test data.

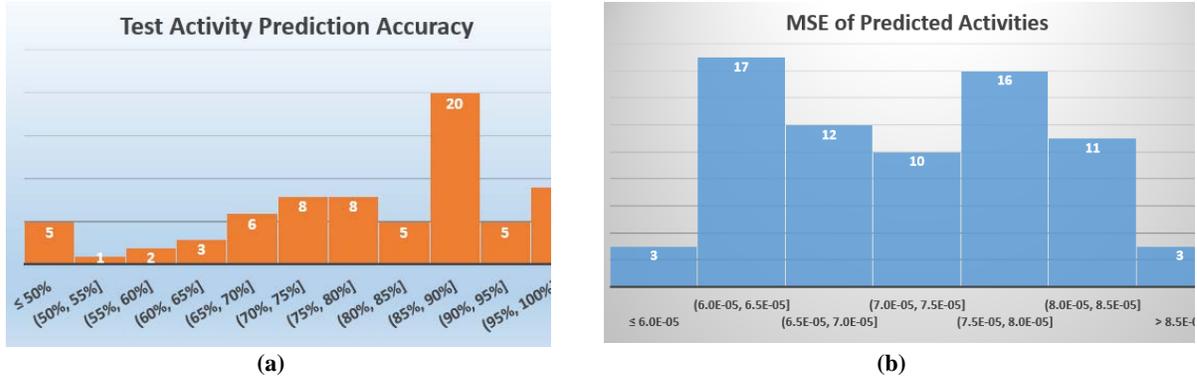

**Figure 4-** (a) The histogram of the test accuracy of exercise name prediction for subject $i$ when subject $i$ is left out during the training; (b)The histogram of the test accuracy of exercise vector representation prediction for subject $i$ when subject $i$ is left out during the training. The accuracy is calculated using MSE between $v_i$ and $\hat{v}_i$.

The second RNN predicts the probability of success for predicted exercises. We use precision to evaluate the performance of the second RNN. Figure 5.a shows the histogram of the precision values for the same experiment described above—leaving out subject $i$ during the training. Figure 5b shows the Brier score of the prediction of the probability of success using RNN and ACT-R models for subject $i$ when subject $i$ is left out during the training. Overall, the RNN model is more accurate in the prediction, expect for five subjects.

Note that the training in all models—both RNNs—are performed on all subjects except for the test subject. Thus, the trained models are able to predict activities for each subject without any personalization. After a certain time and collecting enough data for each subject, we will be able to fine-tune the proposed algorithm for each subject and improve the performance significantly.

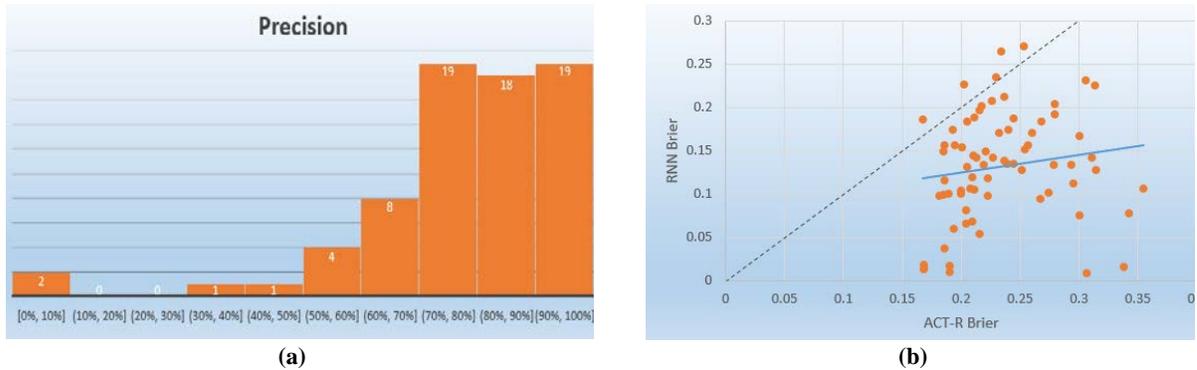

**Figure 5-** (a) The histogram of the precision values for the success prediction using the second RNN model for subject $i$ when subject $i$ is left out during the training; (b) The Brier score for predicting the probability fo success using RNN and ACT-R models.

## 5. Conclusion

The dual-RNN system for recommending workout exercises along with predicting individual success rates achieves high accuracy for individuals from whom we do not have any training data. We validated this achievement by training the proposed model on a set of users and testing on a new set of test users. Even though our time series analysis of the workout data showed short-term dependency, RNN with LSTM units can be deployed for datasets with long term interdependencies.

Similar to many existing recommendation systems, our algorithm suffers when the number of users is small or when the systems are deployed without any prior history of the user. This issue of predictive accuracy is especially problematic in domains where there is a significant downside to making prediction errors. mHealth and behavioral medicine in general are surely examples of such domains. One would not want an algorithm to repeatedly recommend exercises that a person would not or could not perform. The user would likely abandon the app altogether, and the algorithm would never collect enough data to improve. This problem can, of course, be remedied by using hand-crafted recommendation algorithms (such as the DStress example), or digital platforms that use human-to-human counselling as part of the bootstrapping process.

Another approach might involve combinations of explanatory computational models such as ACT-R and machine learning approaches such as the dual-RNN system presented here. With a small amount of engineering and zero parameter estimation, ACT-R was able to predict individual-level exercise success rates "out of the box." Such models can provide initial predictive capabilities and explanations of the psychological mechanisms in play [5]. As even small datasets, from small numbers of participants are collected, we can see that machine learning approaches can overtake the predictive accuracy of those initial psychological models. As more data are collected, one can imagine that patterns learned by machine learning could be used to refine and expand the explanatory mechanisms embodied in the computational cognitive architecture, thus improving both prediction and explanation in tandem.

## ACKNOWLEDGMENTS

Research reported in this paper was supported by the National Institute on Aging of the National Institutes of Health under award number R01AG053163.